\documentclass[times]{article}
\usepackage{geometry}

\usepackage{setspace}
\usepackage{amssymb}
\usepackage{amsmath,graphicx}
\usepackage{color,soul}
\usepackage{bm}
\newcommand{\floor}[1]{\lfloor #1 \rfloor}
\usepackage{tikz}
\def\checkmark{\tikz\fill[scale=0.4](0,.35) -- (.25,0) -- (1,.7) -- (.25,.15) -- cycle;}

\usepackage{lipsum}

\newcommand\blfootnote[1]{%
	\begingroup
	\renewcommand\thefootnote{}\footnote{#1}%
	\addtocounter{footnote}{-1}%
	\endgroup
}

\begin{document}

\date{}

\title{Temporal Logistic Neural Bag-of-Features for Financial Time series Forecasting leveraging Limit Order Book Data}

\author{	Nikolaos Passalis, Anastasios Tefas, Juho Kanniainen, Moncef Gabbouj\\ and Alexandros Iosifidis
	\blfootnote{Nikolaos Passalis, Juho Kanniainen and Moncef Gabbouj are with the Faculty of Information Technology and Communication, Tampere University, Finland. Anastasios Tefas is with the School of Informatics, Aristotle University of Thessaloniki, Greece.  Alexandros Iosifidis is with the Department of Engineering, Electrical and Computer Engineering, Aarhus University, Denmark.  E-mail: nikolaos.passalis@tuni.fi, tefas@csd.auth.gr, juho.kanniainen@tuni.fi, moncef.gabbouj@tuni.fi, alexandros.iosifidis@eng.au.dk
}}

\maketitle
\begin{abstract}
Time series forecasting is a crucial component of many important applications, ranging from forecasting the stock markets to energy load prediction. The high-dimensionality, velocity and variety of the data collected in these applications pose significant and unique challenges that must be carefully addressed for each of them. In this work, a novel Temporal Logistic Neural Bag-of-Features approach, that can be used to tackle these challenges, is proposed. The proposed method can be effectively combined with deep neural networks, leading to powerful deep learning models for time series analysis. However, combining existing BoF formulations with deep feature extractors pose significant challenges: the distribution of the input features is not stationary, tuning the hyper-parameters of the model can be especially difficult  and the normalizations involved in the BoF model can cause significant instabilities during the training process. The proposed method is capable of overcoming these limitations by a employing a novel adaptive scaling mechanism and replacing the classical Gaussian-based density estimation involved in the regular BoF model with a logistic kernel. The effectiveness of the proposed approach is demonstrated using extensive experiments on a large-scale financial time series dataset that consists of more than 4 million limit orders.
\end{abstract}

\section{Introduction}
\label{section:intro}

Time series forecasting is a crucial component of many important applications, ranging from predicting the behavior of financial markets~\cite{cao2003support}, to accurate energy load prediction~\cite{hippert2001neural}. Even though the large amount of data that can be nowadays collected from these domains provide an unprecedented opportunity for applying powerful deep learning (DL) methods~\cite{langkvist2014review, wang2017interactive, lecun2015deep}, the high-dimensionality, velocity and variety of such data also pose significant and unique challenges that must be carefully addressed for each application. To this end, many methods have been proposed to analyze and forecast time series data. For example, traditional approaches employ adaptive distance metrics, such as Dynamic Time Wrapping~\cite{caillault2017dynamic}, to tackle these kind of tasks. However, with the advent of DL the interest is gradually shifting toward using neural network-based methods, including recurrent and convolutional architectures~\cite{lipton2015learning, cui2016multi}, that seem to be more effective for handling such kind of data. It is worth noting that other approaches for time series analysis also exist, such as using the Bag-of-Features model (BoF)~\cite{sivic2003video}. The BoF model was  recently adapted toward efficiently processing large amounts of complex and high-dimensional time series~\cite{baydogan2013bag, bailly2015bag, passalis2018temporal}, due its ability to analyze objects that consist of a varying number of features, as well as withstanding distribution shifts better than competitive methods~\cite{passalis2017learning}.

The Bag-of-Features model (BoF) involves the following pipeline~\cite{sivic2003video}: a) Several feature vectors are extracted from each input object, e.g., an image or time series. This step is called \textit{feature extraction} and allows for forming the \textit{feature space}, where each object is represented as a set of feature vectors. b) A set of representative feature vectors (also called \textit{codewords}) are learned and used to quantize the extracted feature vectors. This step is called \textit{dictionary learning}, while the learned codewords form the \textit{dictionary} (which also called \textit{codebook}). c) The quantized feature vectors are aggregated in order to extract a constant length representation that describes the semantic content/temporal behavior/etc. of each input object.

The BoF model is able to successfully handle objects of various lengths, providing an important advantage over other methods, since it allows for efficiently extracting a constant length representation of time series regardless their actual length. Indeed, the ability of the BoF-based model to tackle several time series analysis tasks have been demonstrated in the literature~\cite{baydogan2013bag, bailly2015bag, passalis2018temporal}. However, these approaches mainly employ shallow models that use simple hand-crafted features, instead of using more powerful deep feature extraction layers for extracting higher level features that can better model the dynamics of a time series~\cite{cui2016multi, tsantekidis2017using}. In this work we argue that combining the BoF model with such architectures can significantly improve the performance of time series forecasting algorithms, since the BoF model allows for dealing with time series of arbitrary lengths and withstanding mild distribution shifts, while using  deep feature extractors, such as recurrent and convolutional layers, allows for taking into account the more detailed temporal dynamics.

\begin{figure*}[ht!]
	\centering
	\includegraphics[width=0.99\linewidth]{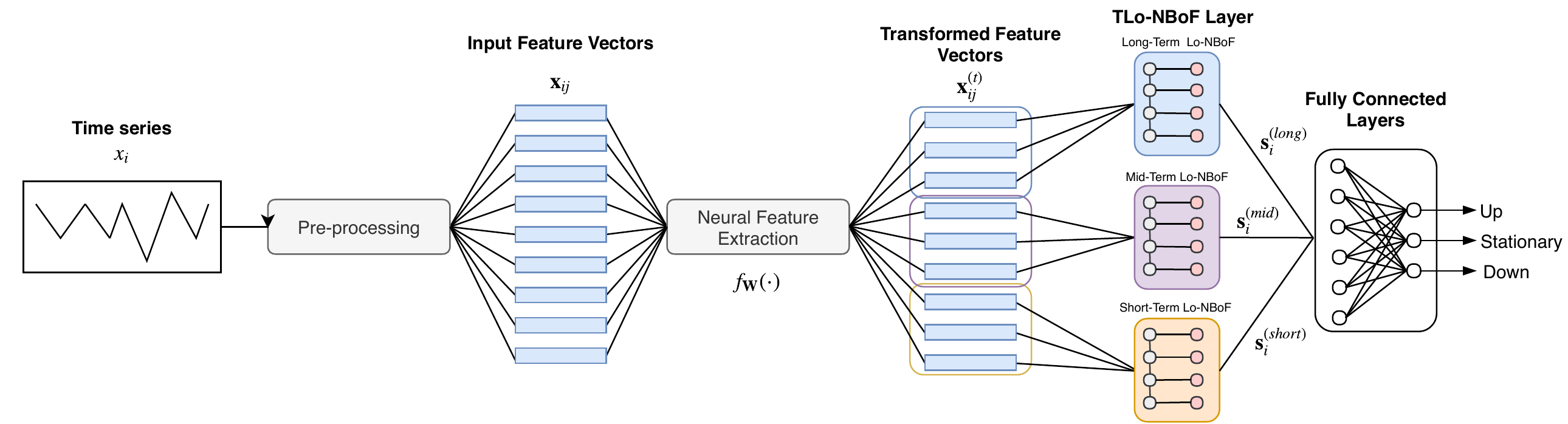}
	\caption{The proposed Temporal Logistic Neural BoF (TLo-NBoF) architecture for time series forecasting.  }
	\label{fig:architecture}
\end{figure*}

The main contribution of this work is the proposal of a novel logistic formulation of the Bag-of-Features model that is adapted toward the needs of time series forecasting and can be effectively combined with deep neural networks. The proposed method is indeed capable of combining the advantages of the BoF model with the enormous learning capacity of deep learning models, allowing for developing powerful forecasting models. However, combining existing BoF formulations with deep feature extractors pose significant challenges: the distribution of the input features is not stationary, tuning the hyper-parameters of the model can be especially difficult  and the normalizations involved in the BoF formulation can cause significant instabilities during the training process. The later was found to be the main cause for the difficulties in training deep models that employ BoF layers and it is addressed by proposing an appropriate adaptive scaling approach. Furthermore, the classical Gaussian-based density estimation involved in the regular BoF model is replaced by a logistic kernel, following a probabilistic formulation of the BoF model~\cite{bhattacharya2011probabilistic}, allowing for further improving the performance of the model and simplifying the implementation without requiring any sophisticated initialization scheme or careful finetuning of any hyper-parameter.  Furthermore, the proposed method is able to perform fine-grained temporal modeling, as shown in Fig.~\ref{fig:architecture}, where the short-term, mid-term, and long-term behavior of time series are modeled. The proposed method is extensively evaluated using a large-scale financial time series dataset that consists of more than 4 million limit orders.

The rest of the paper is structured as follows. First, the related work is briefly introduced and compared to the proposed approach in Section~\ref{sec:related}. Then, the proposed method is introduced in Section~\ref{sec:methodology}, while the experimental evaluation is provided in Section~\ref{sec:experiments}. Finally, conclusions are drawn in Section~\ref{sec:conclusions}.

\section{Related Work}
\label{sec:related}

This work is mainly related to time series analysis using the BoF model. An increasing number of recent works employ variants of the Bag-of-Features model to perform  time series analysis, e.g., forecasting, retrieval, etc. In~\cite{iosifidis2013multidimensional}, a BoF-based method was proposed for extracting discriminative representations by employing a discriminative objective for optimizing the codebook.  A dictionary learning methods for the BoF model was also utilized in~\cite{passalis-entropy}, in order to learn retrieval-oriented representations. A discriminant BoF approach for learning representations for action recognition was proposed in~\cite{iosifidis2014discriminant}, while a dynemes-based one was introduced in~\cite{iosifidis2013dynamic}. Other more recent approaches further adapt the procedure toward time series analysis, e.g., time series segments of various lengths were used in~\cite{baydogan2013bag}, to allow for efficiently handling warping, while an approach that employs temporal modeling was proposed in~\cite{bailly2015bag}. Quite recently, a neural formulation of the BoF model was used to perform time series analysis~\cite{8081217}, while an extension of this method, that allows for better capturing the temporal dynamics of time series, was introduced in~\cite{passalis2018temporal}.

In contrast with~\cite{passalis2018temporal}, in this work a logistic Neural BoF formulation is used. This allows for training temporal BoF models without using any sophisticated initialization schemes and/or carefully tuning any hyper-parameter, e.g., the initial scaling factor of the kernel function that was employed in~\cite{passalis2018temporal}. Furthermore, in this work, we studied behavior of the BoF model when combined with deep feature extractors and we appropriately designed an adaptive scaling method that allows for the smooth flow of information in deep BoF-based architectures. To the best of our knowledge, this is the first work in which a deep temporal formulation of the BoF model is used with deep feature extraction layers, after appropriately adapting it to the needs of the specific application, demonstrating that it is indeed possible to learn powerful deep learning models for time series analysis that outperform other competitive state-of-the-art methods.

\section{Proposed Method}
\label{sec:methodology}

In this Section, the proposed Temporal Logistic Neural Bag-of-Features formulation (abbreviated as ``TLo-NBoF'' in the rest of this paper) is derived and adapted to the needs of modeling the temporal dynamics of high frequency limit-order book data. Furthermore, as it was already discussed in Section~\ref{section:intro}, directly employing a BoF formulation, e.g., N-BoF~\cite{passalis2017neural}, in deep neural networks requires the careful fine-tuning of several hyper-parameters, e.g., separately adjusting the learning rates per layer, carefully selecting the activation functions and the distribution used for initializing the parameters of the network, etc. In this Section, we delve into these issues, examining some of the reasons for the difficulties that arise when Neural BoF formulations are combined with deep neural networks. Then, the proposed model is appropriately adapted to overcome the aforementioned issues, allowing for directly employing it to learn powerful deep neural network architectures for time series analysis.

\subsection{Temporal Logistic Neural Bag-of-Features}

Let $\mathit{x}_i$ be the $i$-th time series of a collection of $N$ training time series, denoted by $\mathcal{X}= \{\mathit{x}_1, \mathit{x}_2, \dots, \mathit{x}_N\}$. Then, several feature vectors can be extracted from each time series using several different approaches, as proposed in the literature~\cite{iosifidis2013multidimensional, kercheval2015modelling, bailly2015bag, fu2011music}. Perhaps the most straightforward one is to directly use the raw time series data for each time step as a separate vector~\cite{iosifidis2013multidimensional}. Depending on the application, more sophisticated methods have been also proposed. For example, when dealing with financial data, domain knowledge can be used to design and extract more rich features that describe several aspects of the time series, e.g., multiple feature vectors can be extracted from high frequency limit order book data using the approach proposed in~\cite{kercheval2015modelling}. The $j$-th feature vector extracted from the $\mathit{x}_i$ time series is denoted by $\mathbf{x}_{ij} \in \mathbb{R}^D$, where $D$ is the dimensionality of the extracted feature vectors. Each time series can be then described by the set of the extracted feature vectors, i.e., $\mathit{x}_i = \{\mathbf{x}_{i1}, \mathbf{x}_{i2}, \dots, \mathbf{x}_{i{N_i}}\}$, where $N_i$ is the length of the $i$-th time series. Note that different time series might have different lengths, so the proposed method must be able to handle objects that are composed of a varying number of feature vectors.

The extracted features are then transformed using a series of neural transformation layers, as shown in Fig.~\ref{fig:architecture}. The employed neural feature extractor is denoted by $f_{W}(\cdot)$, where $W$ are the parameters (weights) of the feature extraction layers. Any (differential) feature extractor can be used to this end, e.g., convolutional layers~\cite{tsantekidis2017forecasting}, recurrent layers~\cite{tsantekidis2017using, DIXON2018277}, etc. In this work, 1-D convolutional layers are used to extract higher level features that are able to capture the temporal relationships between succeeding feature vectors and, as a result, better model the dynamics of a time series. After these neural feature extraction layers, each time series can be represented by a set of \textit{transformed higher level features}. These features are denoted by $ \mathbf{x}^{(t)}_{ij} = f_{W}(\mathit{x}_i, j) \in \mathbb{R}^{D}$, where $D$ the dimensionality of the transformed features. In this work $D$ equals to the number of filters used in the last convolutional layer, since a convolutional layer is used to transform the input features. Therefore, after the neural feature extraction process, the $i$-th time series is described by $\mathit{x}^{(t)}_i = \{\mathbf{x}_{i1}, \mathbf{x}_{i2}, \dots, \mathbf{x}_{i{N_i}}\}$.

Even though the extracted feature vectors $\mathbf{x}^{(t)}_{ij}$ capture higher level information regarding the dynamics of the time series, it is not possible to directly use them for classification purposes (or any other data analysis task), since they have first to be aggregated into a representation that has constant length and it is invariant to the length of the input time series. The performance and flexibility of the resulting model critically depends on the aggregation method that will be employed. For example, the most straightforward approach is to directly flatten the extracted features into one long vector and then feed this vector into a classifer. However, this approach does not allow for a) handling time series that have different lengths, since the length of the resulting vector depends on the length of each series, and b) severely limits the ability of the model to handle time-wrapping/temporal translations/etc.

To overcome the aforementioned limitation, a Temporal Logistic Neural Bag-of-Features formulation is used to efficiently aggregated the extracted feature vectors. Let $\mathcal{V}=\{\mathbf{v}_1, \mathbf{v}_2, \dots, \mathbf{v}_{N_K} \}$ be a dictionary of $N_K$ codewords that are used to quantized the feature vectors extracted using the neural feature extraction layers. Each codeword is denoted by a vector $\mathbf{v}_k \in \mathbb{R}^{D}$. Traditionally, these codewords are either selected using the k-means algorithm~\cite{jain2010data}, or by directly sampling them from the set of the extracted feature vectors~\cite{bhattacharya2011probabilistic}. However, these approaches cannot be used when the BoF model is combined with deep neural networks, since the input feature distribution is not stationary. Instead, it is constantly shifting requiring the codebook to be updated after each training step. To this end, the codewords are considered part of the trainable parameters of the network and directly learned using the regular back-propagation algorithm (as it will be demonstrated later). However, the aforementioned processes (clustering or random sampling) can be still used for initializing the codewords~\cite{passalis2017neural}.

Assuming that each transformed feature vector is generated independently and identically distributed  from an unknown distribution controlled by the (image-specific) the vector $\mathbf{s}_i = ({s}_{i1}, {s}_{i2}, ..., {s}_{iN_K})$, then the probability of observing a transformed feature vector $\mathbf{x}^{(t)}_{ij}$ given the $i$-th time series can be estimated using Kernel Density Estimation~\cite{silverman2018density,katkovnik2002kernel}:
\begin{equation}
p(\mathbf{x}^{(t)}_{ij} | \mathit{x}_i) = \sum_{k=1}^{N_K} s_{ik} K(\mathbf{x}^{(t)}_{ij}, \mathbf{v}_k),
\end{equation}
where $K(\cdot)$ is a kernel and $s_{ik}$ are the time series specific parameters that control the density estimation. The parameter vector $\mathbf{s}_i$ can be estimated using a maximum likehood estimator:
\begin{equation}
\mathbf{{s}}_i = \arg \max_{\mathbf{s}} \sum_{j=1}^{N_i} \log \left( \sum_{k=1}^{N_K} s_{ik} K(\mathbf{x}^{(t)}_{ij},  \mathbf{v}_k) \right).
\end{equation}
It can be easily derived, as also shown in~\cite{bhattacharya2011probabilistic}, that these parameters can be effectively estimated as:
\begin{equation}
\label{eq:s}
s_{ik} = \frac{1}{N_i} \sum_{j=1}^{N_i} u_{ijk},
\end{equation}
where 
\begin{equation}
\label{eq:u}
u_{ijk} = \frac{K(\mathbf{x}^{(t)}_{ij}, \mathbf{v}_k)}{\sum_{l=1}^{N_K} K(\mathbf{x}^{(t)}_{ij}, \mathbf{v}_l)},
\end{equation}
giving rise to the well known BoF with soft-assignments~\cite{van2008kernel,passalis2017neural}. Therefore, the vector $\mathbf{s}_i$ is a histogram that describes the  behavior of the $i$-th time series and can be used for the subsequent classification tasks, apart from using it for estimating the distribution of the feature vectors for each time series.  Furthermore, as shown in our previous work~\cite{passalis2017neural}, (\ref{eq:u}) can be directly implemented as a normalized RBF layer, while (\ref{eq:s}) can be implemented as a recurrent accumulation layer. Usually a Gaussian kernel is used to calculate the normalized membership vector expressed by (\ref{eq:u}):
\begin{equation}
K(\mathbf{x}, \mathbf{v}) = \frac{1}{\sqrt{2 \pi \sigma} } \exp \left( \frac{- ||\mathbf{x} - \mathbf{v} ||^2_2}{2 \sigma^2} \right)
\end{equation}
where $\mathbf{x}$ is a feature vector, $\mathbf{v}$ is a codeword and $\sigma$ is the width of the kernel. However, as shown in~\cite{passalis2017neural, 8506624}, it is not straightforward to select the appropriate kernel width and the performance of the model critically relies on the selection of this parameter (even when $\sigma$ is optimized during the training process). To overcome this limitation, in this paper we propose replacing the Gaussian kernel with a more well-behaved and easy to use sigmoid (also known as hyperbolic) kernel~\cite{dhillon2004kernel}:
\begin{equation}
K(\mathbf{x}, \mathbf{v}) = \tanh(\alpha  \mathbf{x}^T \mathbf{v} + \beta),
\end{equation}
where $\alpha$ and $\beta$ are the parameters of the kernel (usually set to $\alpha=1$ and $\beta=0$) and $tanh(x) = \frac{e^x - e^{-x}}{e^x + e^{-x}}$. The kernel is also scaled to $0 \dots 1$ to ensure that it is compatible with the quantization process employed by (\ref{eq:u}):
\begin{equation}
K(\mathbf{x}, \mathbf{v}) = \frac{1}{2} \left(\tanh (\alpha  \mathbf{x}^T \mathbf{v} + \beta) + 1 \right)=  \text{sigm} (2\alpha \mathbf{x}^T \mathbf{v} + 2\beta),
\end{equation}
where $sigm(x) = \frac{1}{1+ e^{-x}}$ is the logistic sigmoid function. Using this kernel also allows for avoiding the need for sophisticated initialization schemes based on computationally intensive algorithms, e.g., k-means, allowing for simply randomly initializing the codebook, along with the other parameters of the network.

\begin{figure*}[ht!]
	\begin{center}
	\includegraphics[width=0.49\linewidth]{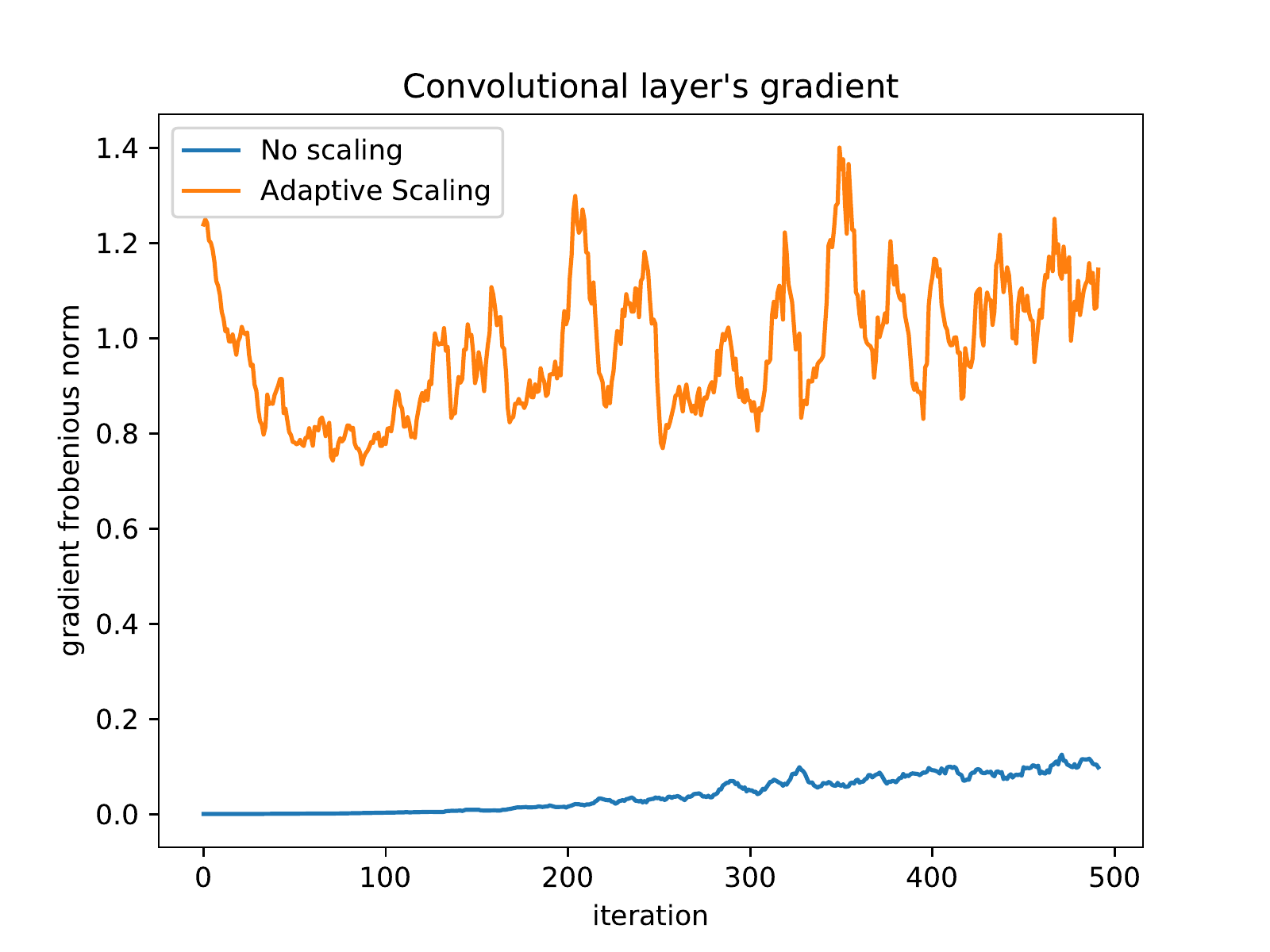}
	\includegraphics[width=0.49\linewidth]{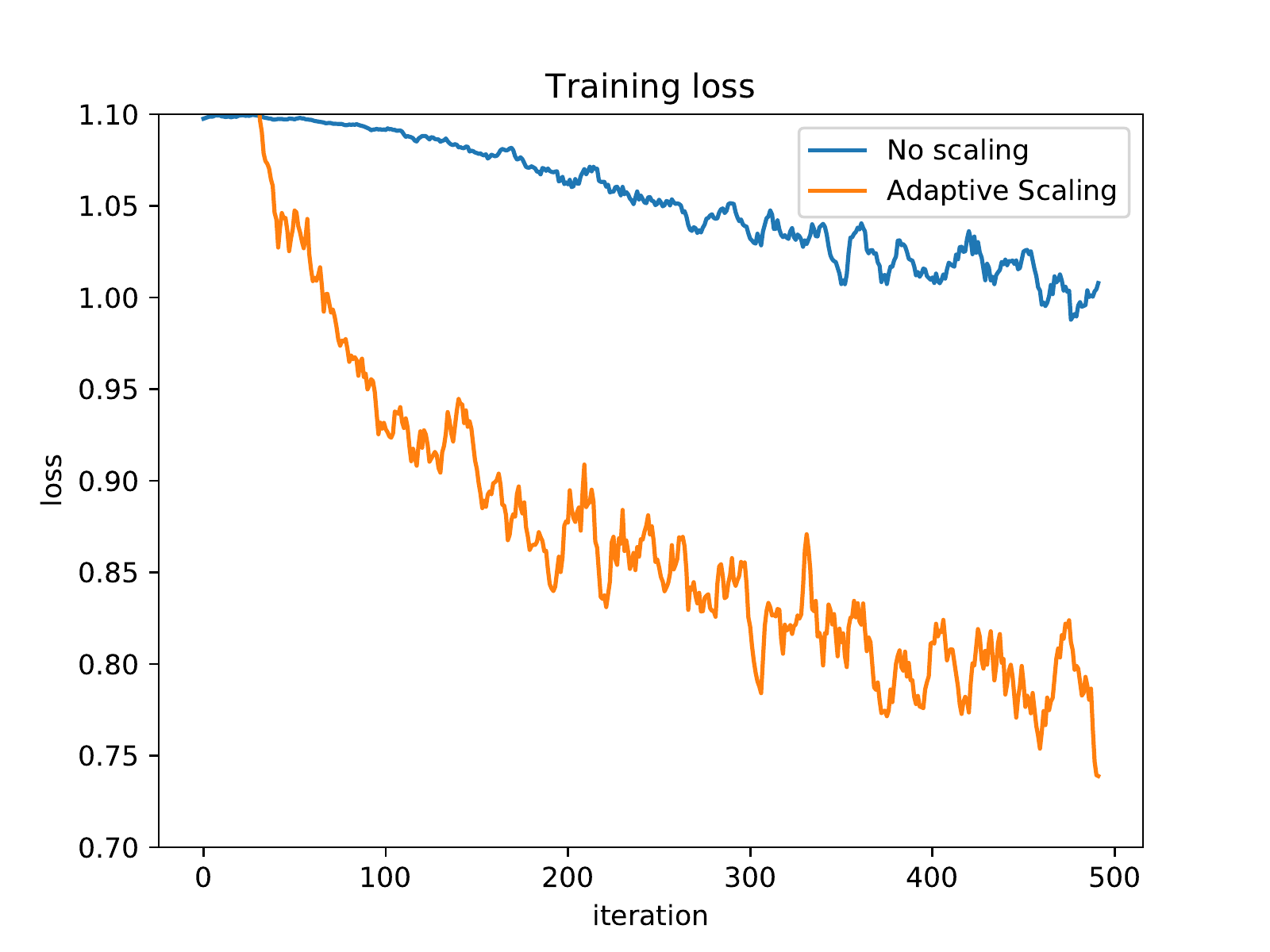}
	\end{center}
	\caption{Effect of using \textit{adaptive scaling}, i.e., allowing the network to adjust the scale of the $\mathbf{s}_i$ and $\mathbf{x}_{ij}$ vectors during the training process, on the gradients of the layers before the proposed TLo-NBoF layer (left figure) and on the loss function (right figure) during the first 500 training iterations. The proposed adaptive scaling approach significantly improves the speed of the convergence of the network.}
	\label{fig:adaptive-scaling}
\end{figure*}

Note that the histogram vector $\mathbf{s}_i$ captures only the overall behavior of the $i$-th time series. In this work, a fine-grained temporal segmentation scheme is also proposed to capture the temporal dynamics of the time series. To this end, the  transformed feature vectors are segmented into $N_T$ temporal regions, as shown in Fig.~\ref{fig:architecture}, to capture the short-term, mid-term, and long-term behavior of the time series ($N_T=3$ temporal regions are used). Therefore, the most recent $\floor{\frac{N_i}{N_T}}$ feature vectors are employed for calculating the short-term histogram $\mathbf{s}^{(short)}_i$, the preceding $\floor{\frac{N_i}{N_T}}$ are used to calculate the mid-term histogram $\mathbf{s}^{(mid)}_i$, while the rest of the feature vectors are used for calculating the long-term histogram $\mathbf{s}^{(long)}_i$. Note that different temporal segmentation schemes can be also applied, i.e., the short-term feature vectors can be fed to both the mid-term and long-term Lo-NBoF models to model the corresponding behavior over longer periods of time. 
Finally, the resulting concatenated vector $\mathbf{s}_i = [\mathbf{s}^{(short)}_i, \mathbf{s}^{(mid)}_i, \mathbf{s}^{(long)}_i] \in \mathbb{R}^{3 N_K}$ is fed to the following fully connected layer, as shown in Fig.~\ref{fig:architecture}.

\subsection{Learning Deep Architectures with Temporal  	Logistic Neural Bag-of-Features}

Even though the previously described architecture can  deal with time series of variable length and capture their fine-grained temporal dynamics, it requires a significant effort in order to tune the appropriate hyper-parameters, e.g., learning rate, initialization, etc., in order to effectively train the resulting architecture. We argue that the main reason for that is the normalizations involved in (\ref{eq:s}) and (\ref{eq:u}). These normalizations, that scales down the $l^1$ norm of the corresponding vectors, prohibit the smooth flow of information, both in the forward and backward propagation. This is illustrated in Fig.~\ref{fig:adaptive-scaling}, where the Frobenius norm of the gradient of the parameters of the first layer of the network are plotted for the first 500 training iterations. Note the extremely small values for the gradients (blue line), that effectively prohibit the gradients from backpropagating to the layers behind the TLo-NBoF model. More than 200 iterations are needed just for starting to slowly update these layers. The harsh scaling involved in (\ref{eq:s}) and (\ref{eq:u}) also reduce the variance of the activation/gradients. However, it is well established that maintaining the same variance for these quantities across the various layers of the network is critical to ensure that the network will be correctly trained, as discussed in detail in~\cite{glorot2010understanding, he2015delving}. Therefore, to overcome these issues, we propose appropriately scaling the $\mathbf{s}_i$ and $\mathbf{u}_{ij}$ vectors as:
\begin{equation}
\label{eq:s1}
s_{ik} = c_s \frac{1}{N_i} \sum_{j=1}^{N_i} u_{ijk},
\end{equation}
and
\begin{equation}
\label{eq:u1}
u_{ijk} = c_u \frac{K(\mathbf{x}^{(t)}_{ij}, \mathbf{v}_k)}{\sum_{l=1}^{N_K} K(\mathbf{x}^{(t)}_{ij}, \mathbf{v}_l)},
\end{equation}
where $c_s$ is initialized to $N_K$, while $c_s$ is initialized to the average number of feature vectors per object. Then, the appropriate values for these two scaling factors are learned during the training process, allowing for easily adjusting the norm of the corresponding vectors to better facilitate the training process. Note that a similar approach have been also used by some activation functions, e.g., PReLU~\cite{he2015delving}, to allow the network to better adjust to the task at hand. Note that the scaling that is involved in~(\ref{eq:s1}) and~(\ref{eq:u1}) still leads to maintaining a constant $l^1$ norm for these vectors, since $c_s$ and $c_u$ are fixed for all the time series that are presented to the network (after the training). This approach, i.e., allowing the network to automatically adjust the norm of the corresponding vectors to allow the smooth flow of information, is called ``\textit{adaptive scaling}'' through this paper. As it can be shown in Fig.~\ref{fig:adaptive-scaling}, where the training process is illustrated during the first 500 training iterations, adaptive scaling can significantly improve the convergence of the network and allows for have more well-behaved gradients on the layers of the network before the employed TLo-NBoF layer.

The resulting architecture, as shown in Fig~\ref{fig:architecture}, can be now directly trained in an end-to-end fashion using gradient descent, i.e.,
\begin{equation}
\Delta(\mathbf{W}_{conv}, \mathbf{V}, \mathbf{W}_{fc}, \mathbf{c}) = \eta (\frac{\partial \mathcal{L}}{\partial \mathbf{W}}, \frac{\partial \mathcal{L}}{\partial \mathbf{V}}, \frac{\partial \mathcal{L}}{\partial \mathbf{W}_{fc}}, \frac{\partial \mathcal{L}}{\partial \mathbf{c}}),
\end{equation}
where $\mathcal{L}$ is the employed loss function, $\mathbf{W}$ denotes the parameters of the neural feature extractor $f_{W}(\cdot)$, $\mathbf{V}=[\mathbf{v}_1, \mathbf{v}_2, \dots, \mathbf{v}_{N_K}]$ denotes the codebook used by the proposed model, $\mathbf{W}_{fc}$ denotes the parameters of the fully connected layers and $\mathbf{c}=(c_u, c_s)$ are the scaling parameters involved in adaptive scaling. The cross-entropy loss is used for all the experiments conducted in this paper, while the same codebook is utilized for all the temporal regions. The Adam algorithm is used to perform the optimization~\cite{kingma2014adam}. The training time series were fed to the network in batches of 128 samples, where each time series was sampled with probability inversely proportional to the frequency of its class. The learning rate was set to $\eta=10^{-4}$, while the networks were trained for 20 epochs. Finally, note that the parameters of the kernel $\alpha$ and $\beta$ can be also optimized during the training process:
\begin{equation}
\Delta(\alpha, \beta) = \eta (\frac{\partial \mathcal{L}}{\partial \alpha}, \frac{\partial \mathcal{L}}{\partial \beta}),
\end{equation}
We refer to this approach, i.e., learning the kernel parameters $\alpha$ and $\beta$, as ``\textit{Kernel Parameter Learning}'', in the rest of this paper.

\begin{table}
	\caption{Architecture of the model employed for  financial time series forecasting}
	\label{table:architecture}
	\begin{center}
		\begin{tabular}{l|c}
			\textbf{Layer} & \textbf{Output size}\\
			\hline
			Input & $N_i \times 144$ \\
			Convolutional (256 filters, kernel size 5)& $N_i \times 256$	\\
			TLo-NBoF ($N_K=256, N_T=3$) & $3\times 256$\\
			Fully Connected (512) & 512\\
			Fully Connected (3) & 3
			
		\end{tabular}
	\end{center}
\end{table}

\begin{table*}[ht!]
	\caption{Ablation Study (the prediction horizon is set to the next 10 time steps)}
	\label{table:ablation}
	\footnotesize
	\begin{center}
		\begin{tabular}{cccc|cc}
			\textbf{Deep Features} & \textbf{Temp. Model.} &  \textbf{Kernel Param. } & 
			
			\textbf{Adaptive Scaling} & \textbf{Macro-F1} & \textbf{Cohen's} $\bm{\kappa}$ \\
						\hline
			- & - & - & \checkmark* & $ 42.66 \pm 0.28 $ & $ 0.1847 \pm 0.0026 $ \\
			\checkmark & - & - & \checkmark* & $ 46.77 \pm 1.53 $ & $ 0.2219 \pm 0.0230 $\\
			- & \checkmark & - & \checkmark* & $ 50.14 \pm 1.36 $ & $ 0.2686 \pm 0.0179 $\\
			\checkmark & \checkmark & - & \checkmark* & $ 51.65 \pm 0.99 $ & $ 0.2783 \pm 0.0109 $\\
			\hline
			\checkmark & \checkmark & \checkmark & - & $ 50.65 \pm 0.71 $ & $ 0.2603 \pm 0.0119 $\\			
			\checkmark & \checkmark & \checkmark & \checkmark* & $ 53.48 \pm 0.45 $ & $ 0.3013 \pm 0.0075 $\\			
			\checkmark & \checkmark & \checkmark & \checkmark &  $\mathbf{53.54 \pm 0.24}$ & $\mathbf{0.3031 \pm 0.0066}$\\	
\end{tabular}
		
	\end{center}
	\footnotesize {(\checkmark* refers to using the scaling parameters $c_s=N_K$ and $c_u=E[N_i]$, but not adjusting them during the training process)}
\end{table*}

\begin{table*}[ht!]
	\caption{Evaluation results using the FI-2010 dataset}
	\label{table:res-main}
	\footnotesize
	\begin{center}
		\begin{tabular}{l|ccccc}
			\textbf{Method} \hspace{12em}  &  \textbf{Macro Precision} &  \textbf{Macro Recall } &  \textbf{Macro F1 score} & \textbf{Cohen's} $\bm{\kappa}$ \\
			\hline
			MLP~\cite{passalis2018temporal} &   $40.20 \pm 0.50$ & $56.25 \pm 2.20$ & $36.91 \pm 1.81 $ & $0.1281 \pm 0.0137 $ \\
			BoF~\cite{passalis2018temporal} &   $39.26 \pm 0.94$ & $51.44 \pm 2.53$ & $36.28 \pm 2.85 $ & $0.1182 \pm 0.0246 $ \\
			N-BoF~\cite{passalis2018temporal} &   $42.28 \pm 0.87$ & $61.41 \pm 3.68$ & $41.63 \pm 1.90 $ & $0.1724 \pm 0.0212 $ \\
			T-BoF~\cite{passalis2018temporal} &   ${43.85 \pm 1.11}$ & ${66.66 \pm 3.40}$ & ${43.96 \pm 1.59}$ & ${0.1992 \pm 0.0201}$ \\
			WMTR~\cite{tran2017tensor} & $46.25 \pm \text{\textit{N/A}}$ &  $51.29 \pm \text{\textit{N/A}}$ & $ 47.87 \pm \text{\textit{N/A}}$ & $\text{\textit{N/A}}$\\
			\hline
			CNN (256 filters) &   $ 44.69 \pm 1.13$ & $ 58.70 \pm 1.85$ & $ 47.19 \pm 1.69 $ & $ 0.2192 \pm 0.0235 $\\
			LSTM (256 neurons) &   $ 47.63 \pm 2.25$ & $ 52.60 \pm 2.74$ & $ 49.51 \pm 2.43 $ & $ 0.2395 \pm 0.0388 $ \\
			GRU (256 neurons) &  $ 47.70 \pm 2.09$ & $ 56.76 \pm 2.99$ & $ 50.55 \pm 2.33 $ & $ 0.2560 \pm 0.0364 $ \\
			TLo-NBoF (proposed) & $ \mathbf{50.20 \pm 2.22}$ & $\mathbf{58.19 \pm 2.13}$ & $ \mathbf{52.98 \pm 2.37}$ & $\mathbf{0.2900 \pm 0.0361}$\\

		\end{tabular}
	\end{center}
\end{table*}

\section{Experimental Evaluation}
\label{sec:experiments}

The proposed method was evaluated using a large-scale limit order book dataset~\cite{ntakaris2018mid, nousi2018machine}. The employed dataset consists of high frequency limit order book data collected from 5 Finish companies traded in the Helsinki Exchange (operated by Nasdaq Nordic). The 10 highest and lower ask order prices were collected for each time step, while data were collected over a period of 10 business days (1st June 2010 to 14th June 2010). A total of 4.5~million limit orders were gathered and processed according to the pre-processing and feature extraction pipeline proposed in~\cite{kercheval2015modelling}. Thus, a total number of 453,975 144-dimensional feature vectors were extracted.

The anchored evaluation setup proposed in~\cite{tomasini2011trading} was used for the evaluation: The time series that were extracted from the first day were used to train the model, while the data from the second day were employed for the evaluation. Then, the first two days were used for the training and the next day was used for the evaluation, etc. This process was repeated 9 times. For all the evaluated metrics  (macro-precision, macro-recall, macro-F1 and Cohen's $\kappa$~\cite{cohen1960coefficient}), the mean and standard deviation are reported.  The direction of the average mid price (up, stationary or down) after 10  time steps was predicted. A stock was considered to be stationary if the change in the mid price was less than to $0.01\%$.

For each time step a time series that consists of the last 15 feature vectors was compiled. The time series was segmented into $N_T=3$ temporal regions, each consisting of 5 feature vectors. The detailed architecture of the employed network is shown in Table~\ref{table:architecture}. First, a convolutional feature extractor with 256 filters (kernel size was set to 5) was used. Then, the transformed feature vectors were fed to a TLo-NBoF layer with 256 codewords. Finally, the temporal histograms extracted from the TLo-NBoF layer were fed into two fully connected layers responsible for predicting the future behavior of the price for the given time series. The ReLU function was used both for the convolutional feature extractor and the first fully connected layer. Note that the TLo-NBoF layer can be directly implemented by using 1D convolutions with kernel size 1, setting the weights of the convolution equal to the codebook and then using the appropriate activation and scaling layers.

First, the effect of the different parts of the proposed architecture are evaluated in the ablation study provided in Table~\ref{table:ablation}. The last three days of the training data were used for performing the ablation study. The first line refers to using a plain Lo-NBoF layer with the appropriate scaling ($c_s$ and $c_u$) to ensure that the model will be successfully trained. Then, adding a convolutional feature extractor (``Deep Features'') improves the Cohen's $\kappa$ by 20\%, while using temporal modeling, i.e., three separate histograms that describe the short-term, mid-term and long-term behavior, improves the Cohen's $\kappa$ by over 40\%. Combining the deep feature extractor with the temporal modeling further improves the performance of the proposed model. Learning the parameters of the kernel (Kernel Parameter Learning) further boost the metrics. Finally note that using and learning the scaling parameters $c_s$ and $c_u$ is crucial for successfully training the network, since without them the performance of the models is reduced, e.g., Cohen's $\kappa$ is reduced by over 14\%.

The proposed method is also compared to various other baselines proposed in the literature~\cite{tran2017tensor,passalis2018temporal}, as well as other powerful convolutional and recurrent deep learning models. For the CNN baseline the same architecture as the one shown in Table~\ref{table:architecture} was used, but the TLo-NBoF layer was replaced by a Global Average Pooling layer. For the GRU~\cite{jozefowicz2015empirical} and LSTM~\cite{hochreiter1997long} models, the feature extraction layers and TLo-NBoF layer were replaced by the appropriate recurrent model. The final state of these models was used for performing the classification. The proposed method significantly improves the performance metrics over both the plain Temporal BoF model~\cite{passalis2018temporal} and the more powerful recurrent and convolutional architectures (the performance is improved by over 13\% over the next best performing model according to the $\kappa$ metric).

\section{Conclusions}
\label{sec:conclusions}

In this paper, a novel logistic formulation of the Neural Bag-of-Features model was proposed and appropriately adapted in order to be efficiently used with deep feature extractors. In this way, the proposed method can effectively combine the advantages of the BoF model with the great learning capacity of DL models, leading to powerful models for time series analysis. The employed fully differential logistic formulation of the BoF model, together with the proposed adaptive scaling mechanism, allows for directly training the resulting architecture in an end-to-end fashion. Furthermore, the proposed method is capable of modeling the behavior of time series at various temporal levels. Finally, the proposed method was extensively evaluated  using a large-scale financial time series dataset that consists of more than 4 million limit orders and it was demonstrated that it performs better than other competitive baseline and state-of-the-art methods.
\\\\
\textbf{Acknowledgments:}
The research leading to these results has received fund-
ing from the H2020 Project BigDataFinance MSCA-ITN-ETN 675044 (http://bigdatafinance.eu), Training for Big Data in Financial Research and Risk Management.

\bibliographystyle{plain}
\bibliography{refs}

\begin{thebibliography}{10}

\bibitem{bailly2015bag}
Adeline Bailly, Simon Malinowski, Romain Tavenard, Thomas Guyet, and Laetitia
  Chapel.
\newblock Bag-of-temporal-sift-words for time series classification.
\newblock In {\em ECML/PKDD Workshop on Advanced Analytics and Learning on
  Temporal Data}, 2015.

\bibitem{baydogan2013bag}
Mustafa~Gokce Baydogan, George Runger, and Eugene Tuv.
\newblock A bag-of-features framework to classify time series.
\newblock {\em IEEE Transactions on Pattern Analysis and Machine Intelligence},
  35(11):2796--2802, 2013.

\bibitem{bhattacharya2011probabilistic}
Subhabrata Bhattacharya, Rahul Sukthankar, Rong Jin, and Mubarak Shah.
\newblock A probabilistic representation for efficient large scale visual
  recognition tasks.
\newblock In {\em Proceedings of the IEEE Conference on Computer Vision and
  Pattern Recognition}, pages 2593--2600, 2011.

\bibitem{caillault2017dynamic}
{\'E}milie~Poisson Caillault, Alain Lefebvre, Andr{\'e} Bigand, et~al.
\newblock Dynamic time warping-based imputation for univariate time series
  data.
\newblock {\em Pattern Recognition Letters}, 2017.

\bibitem{cao2003support}
Li-Juan Cao and Francis Eng~Hock Tay.
\newblock Support vector machine with adaptive parameters in financial time
  series forecasting.
\newblock {\em IEEE Transactions on Neural Networks}, 14(6):1506--1518, 2003.

\bibitem{cohen1960coefficient}
Jacob Cohen.
\newblock A coefficient of agreement for nominal scales.
\newblock {\em Educational and psychological measurement}, 20(1):37--46, 1960.

\bibitem{cui2016multi}
Zhicheng Cui, Wenlin Chen, and Yixin Chen.
\newblock Multi-scale convolutional neural networks for time series
  classification.
\newblock {\em arXiv preprint arXiv:1603.06995}, 2016.

\bibitem{dhillon2004kernel}
Inderjit~S Dhillon, Yuqiang Guan, and Brian Kulis.
\newblock Kernel k-means: spectral clustering and normalized cuts.
\newblock In {\em Proceedings of the ACM SIGKDD International Conference on
  Knowledge Discovery and Data Mining}, pages 551--556, 2004.

\bibitem{DIXON2018277}
Matthew Dixon.
\newblock Sequence classification of the limit order book using recurrent
  neural networks.
\newblock {\em Journal of Computational Science}, 24:277 -- 286, 2018.

\bibitem{fu2011music}
Zhouyu Fu, Guojun Lu, Kai~Ming Ting, and Dengsheng Zhang.
\newblock Music classification via the bag-of-features approach.
\newblock {\em Pattern Recognition Letters}, 32(14):1768--1777, 2011.

\bibitem{glorot2010understanding}
Xavier Glorot and Yoshua Bengio.
\newblock Understanding the difficulty of training deep feedforward neural
  networks.
\newblock In {\em Proceedings of the International Conference on Artificial
  Intelligence and Statistics}, pages 249--256, 2010.

\bibitem{he2015delving}
Kaiming He, Xiangyu Zhang, Shaoqing Ren, and Jian Sun.
\newblock Delving deep into rectifiers: Surpassing human-level performance on
  imagenet classification.
\newblock In {\em Proceedings of the IEEE International Conference on Computer
  Vision}, pages 1026--1034, 2015.

\bibitem{hippert2001neural}
Henrique~Steinherz Hippert, Carlos~Eduardo Pedreira, and Reinaldo~Castro Souza.
\newblock Neural networks for short-term load forecasting: A review and
  evaluation.
\newblock {\em IEEE Transactions on Power Systems}, 16(1):44--55, 2001.

\bibitem{hochreiter1997long}
Sepp Hochreiter and J{\"u}rgen Schmidhuber.
\newblock Long short-term memory.
\newblock {\em Neural Computation}, 9(8):1735--1780, 1997.

\bibitem{iosifidis2013dynamic}
Alexandros Iosifidis, Anastasios Tefas, and Ioannis Pitas.
\newblock Dynamic action recognition based on dynemes and extreme learning
  machine.
\newblock {\em Pattern Recognition Letters}, 34(15):1890--1898, 2013.

\bibitem{iosifidis2013multidimensional}
Alexandros Iosifidis, Anastasios Tefas, and Ioannis Pitas.
\newblock Multidimensional sequence classification based on fuzzy distances and
  discriminant analysis.
\newblock {\em IEEE Transactions on Knowledge and Data Engineering},
  25(11):2564--2575, 2013.

\bibitem{iosifidis2014discriminant}
Alexandros Iosifidis, Anastastios Tefas, and Ioannis Pitas.
\newblock Discriminant bag of words based representation for human action
  recognition.
\newblock {\em Pattern Recognition Letters}, 49:185--192, 2014.

\bibitem{jain2010data}
Anil~K Jain.
\newblock Data clustering: 50 years beyond k-means.
\newblock {\em Pattern recognition letters}, 31(8):651--666, 2010.

\bibitem{jozefowicz2015empirical}
Rafal Jozefowicz, Wojciech Zaremba, and Ilya Sutskever.
\newblock An empirical exploration of recurrent network architectures.
\newblock In {\em Proceedings of the International Conference on Machine
  Learning}, pages 2342--2350, 2015.

\bibitem{katkovnik2002kernel}
Vladimir Katkovnik and Ilya Shmulevich.
\newblock Kernel density estimation with adaptive varying window size.
\newblock {\em Pattern Recognition Letters}, 23(14):1641--1648, 2002.

\bibitem{kercheval2015modelling}
Alec~N. Kercheval and Yuan Zhang.
\newblock Modelling high-frequency limit order book dynamics with support
  vector machines.
\newblock {\em Quantitative Finance}, 15(8):1315--1329, 2015.

\bibitem{kingma2014adam}
Diederik~P. Kingma and Jimmy Ba.
\newblock Adam: A method for stochastic optimization.
\newblock In {\em Proceedings of the International Conference on Learning
  Representations}, 2015.

\bibitem{langkvist2014review}
Martin L{\"a}ngkvist, Lars Karlsson, and Amy Loutfi.
\newblock A review of unsupervised feature learning and deep learning for
  time-series modeling.
\newblock {\em Pattern Recognition Letters}, 42:11--24, 2014.

\bibitem{lecun2015deep}
Yann LeCun, Yoshua Bengio, and Geoffrey Hinton.
\newblock Deep learning.
\newblock {\em Nature}, 521(7553):436, 2015.

\bibitem{lipton2015learning}
Zachary~C Lipton, David~C Kale, Charles Elkan, and Randall Wetzell.
\newblock Learning to diagnose with lstm recurrent neural networks.
\newblock {\em arXiv preprint 1511.03677}, 2015.

\bibitem{nousi2018machine}
Paraskevi Nousi, Avraam Tsantekidis, Nikolaos Passalis, Adamantios Ntakaris,
  Juho Kanniainen, Anastasios Tefas, Moncef Gabbouj, and Alexandros Iosifidis.
\newblock Machine learning for forecasting mid price movement using limit order
  book data.
\newblock {\em arXiv preprint arXiv:1809.07861}.

\bibitem{ntakaris2018mid}
Adamantios Ntakaris, Martin Magris, Juho Kanniainen, Moncef Gabbouj, and
  Alexandros Iosifidis.
\newblock Benchmark dataset for mid-price prediction of limit order book data.
\newblock {\em arXiv preprint arXiv:1705.03233}, 2017.

\bibitem{passalis-entropy}
Nikolaos Passalis and Anastasios Tefas.
\newblock Entropy optimized feature-based bag-of-words representation for
  information retrieval.
\newblock {\em IEEE Transactions on Knowledge and Data Engineering},
  28(7):1664--1677, 2016.

\bibitem{passalis2017learning}
Nikolaos Passalis and Anastasios Tefas.
\newblock Learning bag-of-features pooling for deep convolutional neural
  networks.
\newblock In {\em Proceedings of the IEEE International Conference on Computer
  Vision}, 2017.

\bibitem{passalis2017neural}
Nikolaos Passalis and Anastasios Tefas.
\newblock Neural bag-of-features learning.
\newblock {\em Pattern Recognition}, 64:277--294, 2017.

\bibitem{8506624}
Nikolaos Passalis and Anastasios Tefas.
\newblock Training lightweight deep convolutional neural networks using
  bag-of-features pooling.
\newblock {\em IEEE Transactions on Neural Networks and Learning Systems},
  pages 1--11, 2018.

\bibitem{passalis2018temporal}
Nikolaos Passalis, Anastasios Tefas, Juho Kanniainen, Moncef Gabbouj, and
  Alexandros Iosifidis.
\newblock Temporal bag-of-features learning for predicting mid price movements
  using high frequency limit order book data.
\newblock {\em IEEE Transactions on Emerging Topics in Computational
  Intelligence}, 2018.

\bibitem{8081217}
Nikolaos Passalis, Avraam Tsantekidis, Anastasios Tefas, Juho Kanniainen,
  Moncef Gabbouj, and Alexandros Iosifidis.
\newblock Time-series classification using neural bag-of-features.
\newblock In {\em Proceedings of the European Signal Processing Conference},
  pages 301--305, 2017.

\bibitem{silverman2018density}
Bernard~W Silverman.
\newblock {\em Density estimation for statistics and data analysis}.
\newblock Routledge, 2018.

\bibitem{sivic2003video}
Josef Sivic, Andrew Zisserman, et~al.
\newblock Video google: A text retrieval approach to object matching in videos.
\newblock In {\em Proceedings of the IEEE International Conference on Computer
  Vision}, volume~2, pages 1470--1477, 2003.

\bibitem{tomasini2011trading}
Emilio Tomasini and Urban Jaekle.
\newblock {\em Trading {S}ystems}.
\newblock Harriman House Limited, Hampshire, UK, 2011.

\bibitem{tran2017tensor}
Dat~Thanh Tran, Martin Magris, Juho Kanniainen, Moncef Gabbouj, and Alexandros
  Iosifidis.
\newblock Tensor representation in high-frequency financial data for price
  change prediction.
\newblock In {\em Proceedings of the IEEE Symposium Series on Computational
  Intelligence}, pages 1--7, 2017.

\bibitem{tsantekidis2017forecasting}
Avraam Tsantekidis, Nikolaos Passalis, Anastasios Tefas, Juho Kanniainen,
  Moncef Gabbouj, and Alexandros Iosifidis.
\newblock Forecasting stock prices from the limit order book using
  convolutional neural networks.
\newblock In {\em Proceedings of the IEEE Conference on Business Informatics
  (CBI)}, pages 7--12, 2017.

\bibitem{tsantekidis2017using}
Avraam Tsantekidis, Nikolaos Passalis, Anastasios Tefas, Juho Kanniainen,
  Moncef Gabbouj, and Alexandros Iosifidis.
\newblock Using deep learning to detect price change indications in financial
  markets.
\newblock In {\em Proceedings of the European Signal Processing Conference},
  pages 2511--2515, 2017.

\bibitem{van2008kernel}
Jan~C Van~Gemert, Jan-Mark Geusebroek, Cor~J Veenman, and Arnold~WM Smeulders.
\newblock Kernel codebooks for scene categorization.
\newblock In {\em Proceedings of the European Conference on Computer Vision},
  pages 696--709, 2008.

\bibitem{wang2017interactive}
Yi~Wang, Zhiming Luo, and Pierre-Marc Jodoin.
\newblock Interactive deep learning method for segmenting moving objects.
\newblock {\em Pattern Recognition Letters}, 96:66--75, 2017.

\end{thebibliography}

\end{document}